\documentclass{article}
\usepackage{spconf,graphicx}
\usepackage{amsthm,amsmath,amssymb,amsfonts,bbm}
\usepackage{multirow}       
\usepackage{subcaption}     
\usepackage{xcolor}          
\usepackage[numbers]{natbib}
\usepackage[utf8]{inputenc}
\usepackage[T5,T1]{fontenc} 
\usepackage{ulem} 
\DeclareMathOperator{\arcsinh}{arcsinh}

\newcommand{\remove}{\textcolor[rgb]{1.,.1,.1}} 
\renewcommand{\remove}[1]{}

\title{Multi-scale gridded Gabor attention for cirrus segmentation}
\threeauthors{F. Richards, X. Xie}{Computer Science dpt,\\Swansea University\\Swansea, UK}{A. Paiement}{Université de Toulon,\\Aix Marseille Univ,\\CNRS, LIS,\\Marseille, France}{E. Sola, P.-A. Duc}{Université de Strasbourg,\\CNRS, Observatoire\\astronomique de Strasbourg,\\Strasbourg, France}
\begin{document}
%
\maketitle
%
\begin{abstract}
In this paper, we address the challenge of segmenting global contaminants in large images. 
The precise delineation of such structures requires ample global context alongside understanding of textural patterns. CNNs specialise in the latter, though their ability to generate global features is limited. Attention measures long range dependencies in images, capturing global context, though at a large computational cost. We propose a gridded attention mechanism to address this limitation, greatly increasing efficiency by processing multi-scale features into smaller tiles. We also enhance the attention mechanism for increased sensitivity to texture orientation, by measuring correlations across features dependent on different orientations, in addition to channel and positional attention. We present results on a new dataset of astronomical images, where the task is segmenting large contaminating dust clouds.
\end{abstract}
\section{Introduction}
\label{sec:intro}

Global context is vital in vision: scenes are understood through key descriptive regions, such as grass or sky, as well as through objects. This is especially relevant when processing contaminants covering large regions, such as clouds in remote sensing \cite{Guo2020CDnetV2:Coexistence} and solar imaging, \cite{Feng2020SolarNet:Forecasting}, 
or dust clouds in deep sky imaging \cite{Richards2020LearnableRobustness}.
Multi-scale (MS) CNNs were proposed to increase global context, e.g. \cite{Zhao2017PyramidNetwork}, though context in convolutions remains limited to the final convolutional layers, and achieving larger receptive fields requires downscaling.

Attention proposes to model long range dependencies between feature positions and channels \cite{Vaswani2017AttentionNeed, Fu2019DualSegmentation}.
While attention has been effective in capturing global context, it has a huge computational footprint. Positional attention has squared complexity in relation to image size, which is barely manageable on popular datasets with modern GPUs. This cost of attention can be reduced by downscaling features. Sacrificing texture for gained context is not a worthwhile compromise for some vision tasks, as severe downscaling significantly erodes local textures often to the detriment of model performance.

Orientational information is also a valuable discriminator in identifying classes of objects. Textures are intrinsically composed of orientation dependent patterns, and thus understanding of orientations has been shown to increase performance on a variety of segmentation tasks e.g. \cite{Richards2020LearnableRobustness}.

\begin{figure}
    \centering
    \includegraphics[width=.734\linewidth]{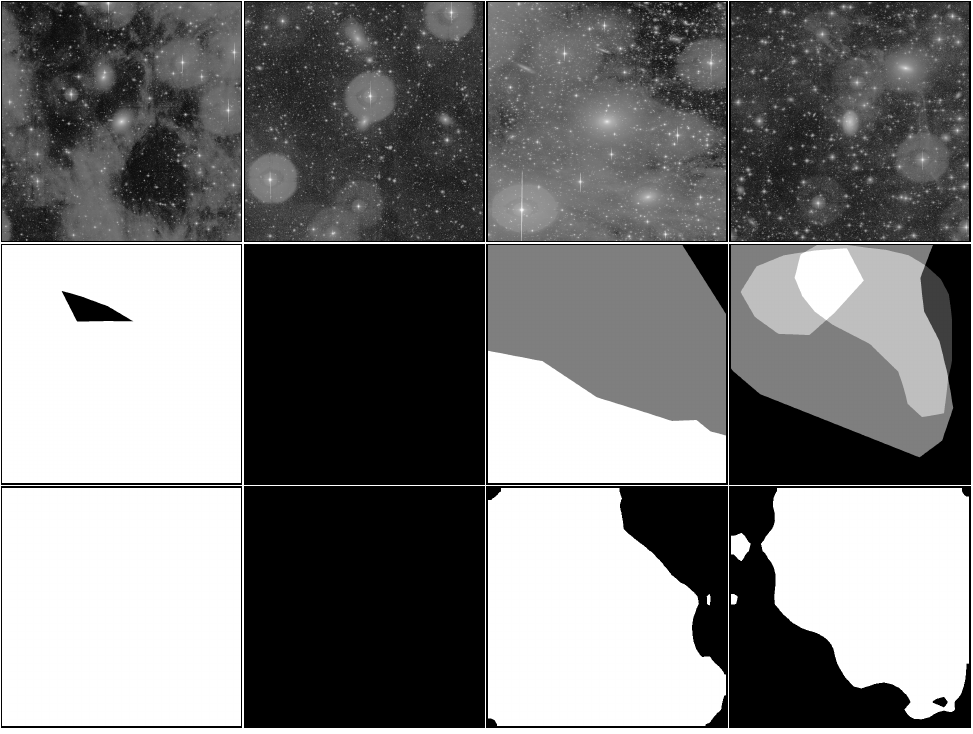}
    \caption{Cirrus dust of various strengths (top), with uncertain annotations (middle) and predictions (bottom).} 
    \label{fig:cirrus_examples_ch5}
\end{figure}

In this paper, we investigate segmentation of cirrus contamination: dust clouds in the foreground of astronomical images. Cirrus pollution can be difficult to spot, ranging from a slight change in background intensity to total occlusion of galaxies, as shown in Fig.~\ref{fig:cirrus_examples_ch5}.
This makes cirrus localisation challenging, leading to frequent disagreements in expert annotations. Background intensity levels vary even in clean regions of images, thus it is necessary to consider the entire image (>5000 px$^2$) to maximise global contextual information. Local textural patterns are also necessary discriminating properties of cirrus, which presents as a wispy texture often with filamentary structures sharing a common orientation.

We propose a gridded MS architecture (Section \ref{subsec:gridded}) that addresses the attention's high compute. Furthermore, it introduces an MS aspect with no added cost, while MS tends to be computationally expensive.
We divide features of different scales into tiles of smaller constant size. 
Positional and channel attention is computed on these tiles to assess both local and global context in an efficient manner, before reassembling tiles into a final attention map. A closely related work is \cite{Sinha2020Multi-scaleSegmentation} where attention is also applied to each scale of MS features, but using whole feature maps in each attention module, resulting in very high computing costs. Additionally, we present a novel attention operator using orientation (Section \ref{subsec:gabor}), for improved sensitivity to textures. We utilise Gabor modulated convolutions \cite{Luan2018GaborNetworks, Richards2020LearnableRobustness} to generate features dependent on different angles. Attention is then computed across these angles, measuring correlations between orientation dependent features. Finally, we propose a loss function to train on annotations that lack a consensus (Section \ref{subsec:loss}), that are a particularity of fuzzy and partially transparent contaminants.

\section{Methodology}
\label{sec:method_ch5}

We present enhancements of attention modules for computationally efficient segmentation of large contaminants. 
These enhancements are compatible with various attention modules (and backbones), and we demonstrate them on \cite{Fu2019DualSegmentation}.

\subsection{Background on Multi-scale Attention}

Accurate identification of contaminants requires comparison to surrounding regions. Attention \cite{Vaswani2017AttentionNeed} has been proposed to analyse contextual relations in images through computing feature correlations. Correlations are scaled based on the strength of given features, prioritising correlations that have a larger effect on model classification. This process can be performed along any internal organisation (axis) of a feature map, allowing different contextual dependencies to be measured. Thus, attention can be computed with respect to position (correlations between regions) and/or with respect to channels (correlations between learned features) \cite{Fu2019DualSegmentation}. We utilise multiple attention operators at different spatial scales to enhance local and global contextual understanding.

An approach for generating MS features is to pool initial features into different sizes and pass them through parallel convolutional layers e.g. \cite{Zhao2017PyramidNetwork}. Feature pyramid approaches \cite{Lin2017FeatureDetection} offer compute savings of constant factor by combining intermediate features at different scales generated by various convolutional layers. They have been used in attention works \cite{Zhao2018Psanet:Parsing, Sinha2020Multi-scaleSegmentation}, with attention applied to each scale apart. \citet{Tao2020HierarchicalSegmentation} generate features for each image scale separately, and observed a performance gain which outweighed computational cost, therefore we use a similar approach although our proposed attentions are compatible with MS features of \cite{Zhao2018Psanet:Parsing, Sinha2020Multi-scaleSegmentation}. 

\subsection{Multi-scale Gridded Attention}
\label{subsec:gridded}

We propose a cost effective method for computing attention over multiple scales. Attention is costly, generally with squared complexity with respect to image size. While this effect can be managed through downscaling and cropping, the former compromises key local texture and the latter compromises key global context. It is therefore desirable to minimise the use of both while introducing a multi-scale analysis.

\begin{figure}
    \centering
    \includegraphics[width=.9\linewidth]{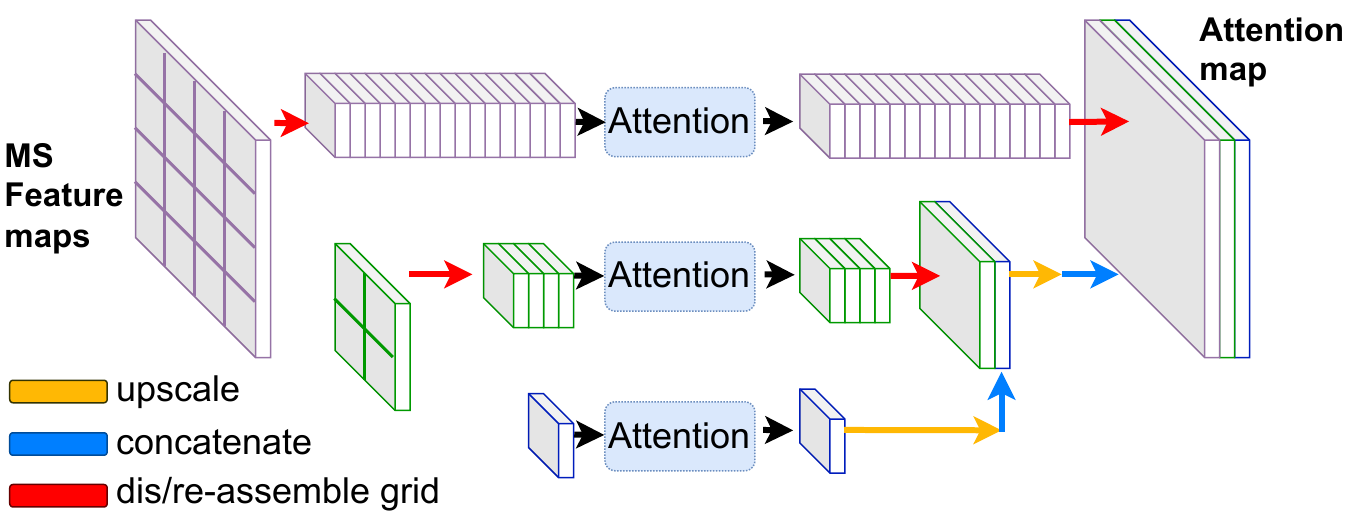}
    \caption{Proposed MS gridded attention.}
    \label{fig:gridded_attention}
\end{figure}

We illustrate gridded attention in Fig.~\ref{fig:gridded_attention}. Our gridded attention module receives input MS feature maps, which we generate as in \cite{Tao2020HierarchicalSegmentation}. Each feature map is concatenated with a fused version of feature maps at all scales, similarly to \cite{Sinha2020Multi-scaleSegmentation}, with the difference that the fused map is produced with same size as the largest feature map (before appropriate rescaling). We divide the feature maps into tiles of consistent smaller size but with multiple underlying spatial scales. Our architecture consists of multiple branches, each handling a spatial scale and comprising of a separate attention module, similar to the multi-branch architecture of \cite{Sinha2020Multi-scaleSegmentation}. In the example of Fig.~\ref{fig:gridded_attention}, attention is computed on 21 tiles with size $\frac{N}{4^2}$, decreasing memory usage by $\frac{21}{256}$. Due to the massive saving on computational resources, the model can inspect both local texture and global context even on large images (e.g. >5000px$^2$ in our case).

\begin{figure}
    \centering
    \includegraphics[width=.86\linewidth]{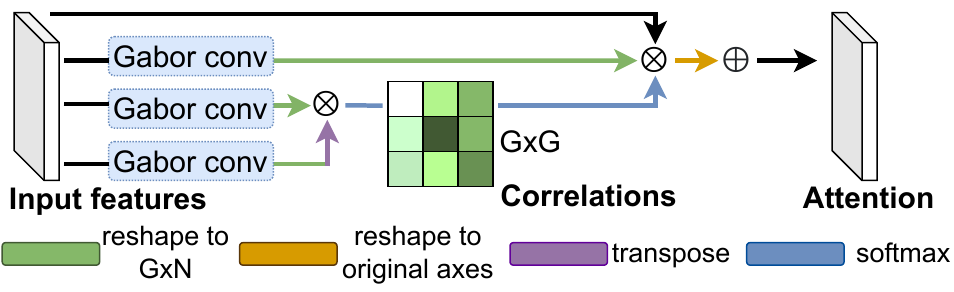}
    \caption{Proposed Gabor attention operator. $G$ is number of modulating Gabor filters, $N$ is the product of other axes. $\bigotimes$ denotes matrix multiplication, and $\bigoplus$ element-wise addition.}
    \label{fig:basic_attention}
\end{figure}

We use spatial scales with a common factor so that rescaling operations can be composed by chaining smaller rescalings. After computing attention on smaller scales, tiles are upscaled to realign resolutions. During upscaling, one upscaling convolutional block is used per scale transition, tying weights across scale branches, increasing parallelisation and inducing overlap between different scales. This is a key difference with other gridded attention methods \cite{Jiang2021Transgan:Up} where lack of tile overlap can cause boundary artefacts.

\subsection{Gabor Attention}
\label{subsec:gabor}

\remove{
We attempt to encode some understanding of a property intrinsic to cirrus clouds, to aid the segmentation model's ability to spot cirrus contamination. Cirrus clouds commonly exhibit streaks with consistent local orientation, which we seek to exploit. Further, textural information alone is not sufficient for reliable annotation due to high variation in background levels, and possible confusion with tidal structures that can appear similar to cirrus locally. To handle these issues, we investigate integrating local and global correlations between orientational features into our segmentation model. In this way, the model can more effectively use the context surrounding uncertain areas to inform the final prediction.
}

We propose a novel attention operator for studying orientational context. Gabor modulated convolutions have been shown to increase performance on oriented textures and rotated samples e.g. \cite{Luan2018GaborNetworks, Richards2020LearnableRobustness}. Such layers multiply convolutional weights by Gabor filters with different rotation parameters to generate orientation dependent features.

As illustrated in Fig.~\ref{fig:basic_attention}, we compute Gabor attention by placing independent Gabor modulations before each reshaping operation and calculate correlations across the axis representing orientations.
This Gabor attention operator may be used in combination with channel and positional operators, creating an attention module with three separate attentions each measuring different dependencies among features. The three attention maps are combined with element-wise summation. We refer to this new attention module, combining positional, channel and Gabor attention, as tri-attention. By measuring correlations across the additional Gabor axis, we are able to study relationships among orientation rich features.



\subsection{Super-Majority Loss function}
\label{subsec:loss}

We propose a super-majority loss (SML) function to train on probabilistic annotations. In this study, targets are generated by four (expert and non-expert) annotators \cite{Sola2022CharacterizationImages}. We create probabilistic targets using a weighted average, where expert annotators are weighted higher. We choose not to use approaches that propose probabilistic networks such as \cite{Hu2019SupervisedAnnotations,Walmsley2020GalaxyLearning} due to our comparatively small number of expert annotators.

Inspired by works on edge detection with CNNs \cite{Liu2017RicherDetection, Xie2015Holistically-nestedDetection}, we separate consensus values into quartiles and conditionally adjust the loss function per quartile. By coarsely dividing probabilities into ranges of confidence, we mitigate against any uncertainty associated with our probabilistic consensuses. We also use focal loss \cite{Lin2017FocalDetection} $L_f$ to encourage the model to focus on difficult examples rather than easy negatives or `clean' pixels which dominate the class balance. Target pixels $y$ with uncertain labels ($0.25 < y < 0.5$) are ignored, whereas super-majority consensuses ($y \geq 0.75$) are prioritised by multiplying with a boosting coefficient $\beta=1.25$. For output $x$:
\begin{align}
    L_{\text{SML}} = \begin{cases}
        \beta \cdot L_f(x, y) & \text{if $y \geq 0.75$}.\\
        L_f(x, y) & \text{if $0.5\leq y < 0.75$}.\\
        0 & \text{if $0.25 < y < 0.5$}.\\
        L_f(x, y) & \text{otherwise}.
    \end{cases} 
\end{align}

\subsection{Final Segmentation Model}


We use the proposed MS gridded and Gabor attention mechanisms in combination, by using the Gabor tri-attention module in the MS gridded framework. We implement an MS attention with three spatial scales. The original feature maps and resulting attention maps are then passed through a fully connected layer to generate segmentation predictions.

As in \cite{Sinha2020Multi-scaleSegmentation}, we generate six segmentation predictions. During training, we consider each segmentation separately and compute six loss terms for all predictions before summing them. The backbone is then explicitly forced to preserve spatial locations of features, relieving the attention module of any realigning effort. For inference, we take the average of the three segmentation computed from the attention maps.

\section{Results}
\label{subsec:data_ch5}

\remove{In this section, we detail techniques we use in combination to craft a robust training strategy for training on annotated LSB images, addressing the data-specific challenges related to this work. First, training data is described. Second, we detail the transfer learning and data augmentation used to mitigate against model overfitting and improve generalisation. Third, we present a pixel rescaling operation which we encode as a network layer with parameters that can be learned in an end to end fashion. Finally, we present a novel loss function for consensuses of annotations and construct the total loss function used for training our network.}

\subsection{Data}

We produced a dataset of 186 images from the MATLAS survey \cite{Duc2020MATLAS:Galaxies} (avg. image size 7000 px$^2$). 25\% of images have cirrus contamination, and in contaminated images, 60\% of pixels contain cirrus.
We use 70\% of samples for training, 15\% for validation and testing. We also run experiments on the subset of contaminated images, and on an additional dataset of 300 synthesised cirrus images (size 512 px$^2$) generated similarly to \cite{Richards2020LearnableRobustness} with some added simulated imaging artefacts for greater realism. In these two additional datasets, class imbalance is eased, allowing the model to focus further on discriminative cirrus features 
and thus more clearly revealing findings during ablation studies. We augment images with geometric transformations: flips, rotations and translations, and apply element-wise Gaussian noise with zero mean and variance of $0.1$. We also pretrain on synthetised cirrus images.

\remove{
We augment images with geometric transformations: flips, rotations and translations, and apply element-wise Gaussian noise with zero mean and variance of $0.1$.
We also 
pretrain on synthetised cirrus images.
}

\subsection{Comparative Analyses on Proposed Techniques}

We verify the effectiveness of the proposed methods individually. For control model, we use non-gridded MS dual attention similar to \cite{Sinha2020Multi-scaleSegmentation} in our MS architecture, though with a simple backbone with four convolutional layers. All networks are trained for 200 epochs with the Adam optimiser \cite{Kingma2014Adam:Optimization} with learning rate $10^{-3}$ and L2 weight regularisation $10^{-7}$. Learning rate is decreased every epoch by a factor of 0.98.  We score models using the intersection over union (IoU) averaged over five splits, and also report standard error across splits. We fix batch size and overall parameter size across modifications to be roughly equal and downscale images to the maximum size our GPUs could accommodate for each network.

\begin{table}[b]
\small
\begin{center}
\begin{tabular}{l|lll}
\hline
               & All images       & Cirrus only      & Synth.       \\ \hline
Control        & $0.469(0.005)$ & $0.814(0.013)$ & $0.831(0.001)$ \\
SML            & $0.483(0.006)$ & $0.822(0.013)$ & $0.830(0.001)$ \\
SML+Gabor      & $0.497(0.007)$ & $0.861(0.015)$ & $0.867(0.002)$ \\
SML+Grid       & $0.542(0.003)$ & $0.886(0.016)$ & $0.844(0.003)$ \\
All  & $\textbf{0.548}(0.003)$ & $\textbf{0.892}(0.013)$ & $0.871(0.002)$ \\ \hline
U-Net \cite{Ronneberger2015U-net:Segmentation} & $0.381(0.128)$ & $0.685(0.096)$ & $0.794(0.001)$ \\
LGCN \cite{Richards2020LearnableRobustness} & $0.414(0.049)$ & $0.741(0.033)$ & $\textbf{0.891}(0.001)$ \\ \hline
\end{tabular}
\caption{Segmentation IoU in format mean(std).}
\label{tab:ablation}
\end{center}
\end{table}

\begin{table}[h]
\small
\begin{center}
\begin{tabular}{lll}
\hline
     & Single & Ensemble \\ \hline
IoU  & 0.745    & \textbf{0.790} \\
Dice & 0.766    & \textbf{0.814} \\ \hline
\end{tabular}
\caption{Proposed cirrus segmentation results} 
\label{tab:bignetwork_results}
\end{center}
\end{table}

\begin{table}[h]
\small
\begin{center}
\begin{tabular}{lllll}
\hline
     & Ours           & \citet{Dev2017Color-BasedCameras} & \citet{Dev2019CloudSegNet:Segmentation} & \citet{Song2020AnDatasets} \\ \hline
IoU   & \textbf{0.90} & 0.69                              & 0.80                                    & 0.86                       \\
Dice  & \textbf{0.95} & 0.82                              & 0.89                                    & 0.92                       \\ \hline
\end{tabular}
\caption{Segmentation scores on the SWIMSEG sky/cloud segmentation dataset.}
\label{tab:swimseg}
\end{center}
\end{table}



We perform an ablation study on proposed attention techniques, shown in Table \ref{tab:ablation}. The effectiveness of the proposed loss is first compared against a focal loss with rounded probabilistic labels \ref{tab:ablation}. On our dataset of real images containing probabilistic annotations, ignoring pixels with uncertain annotation and prioritising very certain pixels is helpful, with the proposed super-majority loss increasing performance. On the synthetic dataset, the probabilities of simulated labels do not correlate well with actual prediction difficulty, and the proposed loss was not able to provide an improvement. We see that the control MS attention model offers a sizeable benefit on real data, indicating that dual attention is well suited to large contaminant segmentation. The proposed MS gridded attention significantly compounds this positive effect, showing that the ability to study long range dependencies across multiple scales is beneficial. The proposed tri-attention increases accuracy on all cirrus segmentation scenarios, indicating that sensitivity to texture orientation is helpful in localising the fuzzy and textured contaminants. 

The two attention techniques benefit from being combined, with further improved results. On real images, when compared against a popular U-Net \cite{Ronneberger2015U-net:Segmentation} baseline and the state-of-the-art cirrus segmentation method LGCN \cite{Richards2020LearnableRobustness}, it outperforms both by a large margin. On synthetic images, LGCN outperforms attention models due to no long range correlations existing within the small 
images.




\subsection{Proposed Cirrus Segmentation}
\label{sec:real_results_ch5}

We construct a final model from the best performing components of the previous section and analyse the predictions. We significantly increase the parameter size of our segmentation model and swap the simple backbone network out for a ResNet-50 \cite{He2016DeepRecognition} network. To account for this larger network, we increase training epochs to 400. We also increase the size of both the training and testing sets by dividing the validation samples between them. Cirrus dust is faint in comparison to bright stars. To compensate for this we add an initial layer to the network implementing $\arcsinh$ scaling, popular in astronomical image processing: $X_s=\arcsinh{(aX+b)}$, where $a, b$ are learned. 
We also use ensemble predictions, where 5 models are trained, and predictions from each model are averaged over to give the final segmentation.
Inference takes an average time of $\sim0.4\text{s}$ per sample on a single GTX 1080 Ti, which could be significantly reduced with optimisation efforts such as model compilation, pruning or half precision weights.


In Table \ref{tab:bignetwork_results} we report both IoU and Dice scores, with Dice being more biased to precision. 
We observe that there is a very large performance increase from models in the comparative analysis, owing to the larger parameter space, additional training samples and longer training period. We see that a single model is outperformed by the model ensemble by a significant margin. Aggregating predictions also appears to handle the class imbalance issue better than the single model, when comparing the predicted distributions of cloud coverage across images against target groundtruth distribution, 
with Kullback-Leibler divergences being respectively 0.40 and 0.07 for the single and ensemble models. Prediction aggregation appears to also increase the gap between IoU and Dice scores, indicating that precision is increased through averaging over predictions. This may be due to the aggregation smoothing over-eager positive predictions from single models within the ensemble. Fig.~\ref{fig:cirrus_examples_ch5} shows some examples of predictions by the best performing ensemble model.

\remove{
\begin{figure}
    \centering
    \includegraphics[width=1.1\linewidth]{Images/chap5/class_hists.pdf}
    \caption{Histograms showing the proportion of target and predicted cirrus coverage across testing images. Left to right: groundtruth, and predictions from the single and ensemble models respectively.}
    \label{fig:bignetwork_cirrus_target_hist}
\end{figure}
}


\remove{
To better understand the characteristics of the trained segmentation models, we turn to a brief qualitative assessment of generated predictions. As shown in Fig.~\ref{fig:NGC2592-NGC6703-NGC7465-UGC04375},
\begin{figure}[t!]
    \centering
    \includegraphics[width=\linewidth]{Images/chap5/NGC2592-NGC6703-NGC7465-UGC04375.pdf}
    \caption{Segmentation predictions on examples with high cirrus coverage. Columns three and four show predictions generated by a single model and an ensemble of models, respectively. Light grey in the prediction map, as in the third row, indicates where the model predicted an uncertain pixel as positive.}
    \label{fig:NGC2592-NGC6703-NGC7465-UGC04375}
\end{figure}
the model deals well with regions entirely contaminated by cirrus dust, with few errors. The inner section of cirrus clouds appears to be reliably predicted, though the model struggles with matching the envelope of contaminated regions. The model handles regions with no cirrus contamination very well, as shown in Fig.~\ref{fig:NGC5846-NGC6017-NGC6547-UGC03960}, 
\begin{figure}[t!]
    \centering
    \includegraphics[width=\linewidth]{Images/chap5/NGC5846-NGC6017-NGC6547-UGC03960.pdf}
    \caption{Segmentation predictions on difficult examples with no cirrus coverage. In this figure, we specifically chose examples with regions that present similarly to cirrus contamination, such as large regions of diffuse light in NGC5846 (first row), or large areas of high background levels in UGC03960 (fourth row).}
    \label{fig:NGC5846-NGC6017-NGC6547-UGC03960}
\end{figure}
even those with high background levels (NGC6017, UGC03960) or large areas of diffuse light (NGC5846). It can also be seen that the ensemble model does appear to increase accuracy in some uncertain scenarios, demonstrating the strength of averaging over multiple predictions. Fig.~\ref{fig:NGC6278-NGC6798-NGC7710-PGC016048} 
\begin{figure}[t!]
    \centering
    \includegraphics[width=\linewidth]{Images/chap5/NGC6278-NGC6798-NGC7710-PGC016048.pdf}
    \caption{The four segmentation predictions with the lowest IoU scores across the testing set. Dark grey in the prediction map, such as in the third row, indicates where the model predicted an uncertain pixel as negative.}
    \label{fig:NGC6278-NGC6798-NGC7710-PGC016048}
\end{figure}
shows the 4 samples with the lowest IoU scores: while the model performs well on most images, there are some examples with poor accuracy. In particular the model appears to struggle with more localised areas of contamination, especially when contamination is close to the boundary where there is a small amount of global context. 
}

\remove{
We report both IoU and Dice scores for model predictions in this section. Dice is more biased to precision rather than recall, so the use of both metrics provides provides a view of the predictive characteristics of our trained models. Fig.~\ref{fig:bignetwork_training_curves} shows the training curves for a training run of the prediction model. We observe that there is a significant difference between training and testing scores, which is expected given the limited sample size of our dataset. Regardless of this, testing performance does trend upwards along with training performance, indicating that the network is not overfitting. We also see that Dice testing scores seem to increase at a higher rate and for longer than IoU scores, showing that as the model is trained further it predicts less false positives than false negatives relative to the true positive predictions.
\begin{figure}[h]
    \centering
    \begin{subfigure}[b]{.48\linewidth}
        \includegraphics[width=\linewidth]{Images/chap5/bignetwork_iou.pdf}
        \caption{IoU.}
        \label{fig:bignetwork_iou}
    \end{subfigure}
    \begin{subfigure}[b]{.48\linewidth}
        \includegraphics[width=\linewidth]{Images/chap5/bignetwork_dice.pdf}
        \caption{Dice.}
        \label{fig:bignetwork_dice}
    \end{subfigure}
    \caption{Training curves (smoothed) for the proposed model showing how IoU and Dice scores change over training epochs on the training and testing sets. We also fit a logarithmic curve to each plot to help illustrate the convergence trend.}
    \label{fig:bignetwork_training_curves}
\end{figure}
}

\subsection{Cloud segmentation in natural images}

We evaluate the proposed methodology on cloud segmentation in natural images from the Singapore Whole sky IMaging SEGmentation database \cite{Dev2017Color-BasedCameras} (SWIMSEG). This task is highly relevant to the proposed methodology: both local and global features are key to good performance on this dataset, as difficult positive regions of light cloud can only be correctly identified based on subtle textural patterns and comparison with surrounding regions. SWIMSEG contains 1013 600px$^2$ images of sky patches and corresponding cloud segmentations. Training, validation and testing sets each contain 861, 101 and 51 samples respectively. The model and training setup used is identical to those of Section~\ref{sec:real_results_ch5}. Results are detailed in Table \ref{tab:swimseg} where we report the segmentation IoU and Dice score on the testing set as in \cite{Dev2017Color-BasedCameras, Song2020AnDatasets, Dev2019CloudSegNet:Segmentation}. The tri-attention model achieves respective IoU and Dice scores of 0.90 and 0.95, representing a significant improvement over state of the art. 

\section{Conclusion}
\label{sec:conclusion_ch5}

We presented a computationally efficient multi-scale attention architecture that is sensitive to texture orientation for segmentation of global contaminants in large images. Efficiency is achieved through a gridded architecture, 
and orientation sensitivity is provided by a new Gabor attention operator.
To address the challenge of uncertain groundtruth labelling, we proposed a simple consensus loss. 
These contributions are combined into a new state-of-the-art model for the segmentation of cirrus contaminants from astronomical images. Reliable performance was achieved with only a small training dataset of images. Our model can process an entire image in one pass with minimal downscaling, meaning that the proposed method can be easily integrated into data processing pipelines for imaging instruments to obtain contamination masks.
In future work it would be interesting, using the methodologies presented in this work, to craft a deep generative model capable of removing global contaminants.

\vfill\pagebreak

\bibliography{references}

\begin{thebibliography}{25}
\providecommand{\natexlab}[1]{#1}
\providecommand{\url}[1]{\texttt{#1}}
\expandafter\ifx\csname urlstyle\endcsname\relax
  \providecommand{\doi}[1]{doi: #1}\else
  \providecommand{\doi}{doi: \begingroup \urlstyle{rm}\Url}\fi

\bibitem[Dev et~al.(2017)Dev, Lee, and Winkler]{Dev2017Color-BasedCameras}
Soumyabrata Dev, Yee~Hui Lee, and Stefan Winkler.
\newblock {Color-Based Segmentation of Sky/Cloud Images From Ground-Based
  Cameras}.
\newblock \emph{IEEE J-STARS}, 10\penalty0 (1):\penalty0 231--242, 2017.

\bibitem[Dev et~al.(2019)Dev, Nautiyal, Lee, and
  Winkler]{Dev2019CloudSegNet:Segmentation}
Soumyabrata Dev, Atul Nautiyal, Yee~Hui Lee, and Stefan Winkler.
\newblock {CloudSegNet: A deep network for nychthemeron cloud image
  segmentation}.
\newblock \emph{IEEE Geoscience and Remote Sensing Letters}, 16\penalty0
  (12):\penalty0 1814--1818, 2019.

\bibitem[Duc(2020)]{Duc2020MATLAS:Galaxies}
Pierre-Alain Duc.
\newblock {MATLAS: a deep exploration of the surroundings of massive early-type
  galaxies}.
\newblock \emph{arXiv preprint arXiv:2007.13874}, 2020.

\bibitem[Feng and Zhang(2020)]{Feng2020SolarNet:Forecasting}
Cong Feng and Jie Zhang.
\newblock {SolarNet: A sky image-based deep convolutional neural network for
  intra-hour solar forecasting}.
\newblock \emph{Solar Energy}, 204:\penalty0 71--78, 2020.

\bibitem[Fu et~al.(2019)Fu, Liu, Tian, Li, Bao, Fang, and
  Lu]{Fu2019DualSegmentation}
Jun Fu, Jing Liu, Haijie Tian, Yong Li, Yongjun Bao, Zhiwei Fang, and Hanqing
  Lu.
\newblock {Dual attention network for scene segmentation}.
\newblock In \emph{CVPR}, pages 3146--3154, 2019.

\bibitem[Guo et~al.(2020)Guo, Yang, Yue, Tan, Hou, and
  Li]{Guo2020CDnetV2:Coexistence}
Jianhua Guo, Jingyu Yang, Huanjing Yue, Hai Tan, Chunping Hou, and Kun Li.
\newblock {CDnetV2: CNN-based cloud detection for remote sensing imagery with
  cloud-snow coexistence}.
\newblock \emph{IEEE Transactions on Geoscience and Remote Sensing},
  59\penalty0 (1):\penalty0 700--713, 2020.

\bibitem[He et~al.(2016)He, Zhang, Ren, and Sun]{He2016DeepRecognition}
Kaiming He, Xiangyu Zhang, Shaoqing Ren, and Jian Sun.
\newblock {Deep residual learning for image recognition}.
\newblock In \emph{CVPR}, pages 770--778, 2016.

\bibitem[Hu et~al.(2019)Hu, Worrall, Knegt, Veeling, Huisman, and
  Welling]{Hu2019SupervisedAnnotations}
Shi Hu, Daniel Worrall, Stefan Knegt, Bas Veeling, Henkjan Huisman, and Max
  Welling.
\newblock {Supervised uncertainty quantification for segmentation with multiple
  annotations}.
\newblock In \emph{MICCAI}, pages 137--145, 2019.

\bibitem[Jiang et~al.(2021)Jiang, Chang, and Wang]{Jiang2021Transgan:Up}
Yifan Jiang, Shiyu Chang, and Zhangyang Wang.
\newblock {Transgan: Two pure transformers can make one strong gan, and that
  can scale up}.
\newblock \emph{Advances in Neural Information Processing Systems}, 34, 2021.

\bibitem[Kingma and Ba(2014)]{Kingma2014Adam:Optimization}
Diederik~P Kingma and Jimmy Ba.
\newblock {Adam: A method for stochastic optimization}.
\newblock In \emph{ICLR}, 2014.

\bibitem[Lin et~al.(2017{\natexlab{a}})Lin, Doll{\'{a}}r, Girshick, He,
  Hariharan, and Belongie]{Lin2017FeatureDetection}
Tsung-Yi Lin, Piotr Doll{\'{a}}r, Ross Girshick, Kaiming He, Bharath Hariharan,
  and Serge Belongie.
\newblock {Feature pyramid networks for object detection}.
\newblock In \emph{ICPR}, pages 2117--2125, 2017{\natexlab{a}}.

\bibitem[Lin et~al.(2017{\natexlab{b}})Lin, Goyal, Girshick, He, and
  Doll{\'{a}}r]{Lin2017FocalDetection}
Tsung-Yi Lin, Priya Goyal, Ross Girshick, Kaiming He, and Piotr Doll{\'{a}}r.
\newblock {Focal Loss for Dense Object Detection}.
\newblock In \emph{ICCV}, pages 2999--3007, 2017{\natexlab{b}}.

\bibitem[Liu et~al.(2017)Liu, Cheng, Hu, Wang, and Bai]{Liu2017RicherDetection}
Yun Liu, Ming-Ming Cheng, Xiaowei Hu, Kai Wang, and Xiang Bai.
\newblock {Richer convolutional features for edge detection}.
\newblock In \emph{CVPR}, pages 3000--3009, 2017.

\bibitem[Luan et~al.(2018)Luan, Chen, Zhang, Han, and
  Liu]{Luan2018GaborNetworks}
Shangzhen Luan, Chen Chen, Baochang Zhang, Jungong Han, and Jianzhuang Liu.
\newblock {Gabor convolutional networks}.
\newblock \emph{IEEE Transactions on Image Processing}, 27\penalty0
  (9):\penalty0 4357--4366, 2018.

\bibitem[Richards et~al.(2020)Richards, Paiement, Xie, Sola, and
  Duc]{Richards2020LearnableRobustness}
Felix Richards, Adeline Paiement, Xianghua Xie, Elisabeth Sola, and
  Pierre-Alain Duc.
\newblock {Learnable Gabor modulated complex-valued networks for orientation
  robustness}.
\newblock \emph{arXiv preprint arXiv:2011.11734}, 2020.

\bibitem[Ronneberger et~al.(2015)Ronneberger, Fischer, and
  Brox]{Ronneberger2015U-net:Segmentation}
Olaf Ronneberger, Philipp Fischer, and Thomas Brox.
\newblock {U-net: Convolutional networks for biomedical image segmentation}.
\newblock In \emph{MICCAI}, pages 234--241, 2015.

\bibitem[Sinha and Dolz(2020)]{Sinha2020Multi-scaleSegmentation}
Ashish Sinha and Jose Dolz.
\newblock {Multi-scale self-guided attention for medical image segmentation}.
\newblock \emph{IEEE J. Biomed. Health Inform.}, 25\penalty0 (1):\penalty0
  121--130, 2020.

\bibitem[Sola et~al.(2022)Sola, Duc, Richards, Paiement, Urbano, Klehammer,
  B{\'{i}}lek, Cuillandre, Gwyn, and
  McConnachie]{Sola2022CharacterizationImages}
Elisabeth Sola, Pierre-Alain Duc, Felix Richards, Adeline Paiement, Mathias
  Urbano, Julie Klehammer, Michal B{\'{i}}lek, Jean-Charles Cuillandre, Stephen
  Gwyn, and Alan McConnachie.
\newblock {Characterization of low surface brightness structures in annotated
  deep images}.
\newblock \emph{Astronomy {\&} Astrophysics}, 662:\penalty0 A124, 2022.

\bibitem[Song et~al.(2020)Song, Cui, and Liu]{Song2020AnDatasets}
Qianqian Song, Zhihui Cui, and Pu~Liu.
\newblock {An Efficient Solution for Semantic Segmentation of Three
  Ground-based Cloud Datasets}.
\newblock \emph{Earth and Space Science}, 7\penalty0 (4), 2020.

\bibitem[Tao et~al.(2020)Tao, Sapra, and
  Catanzaro]{Tao2020HierarchicalSegmentation}
Andrew Tao, Karan Sapra, and Bryan Catanzaro.
\newblock {Hierarchical multi-scale attention for semantic segmentation}.
\newblock \emph{arXiv preprint arXiv:2005.10821}, 2020.

\bibitem[Vaswani et~al.(2017)Vaswani, Shazeer, Parmar, Uszkoreit, Jones, Gomez,
  Kaiser, and Polosukhin]{Vaswani2017AttentionNeed}
Ashish Vaswani, Noam Shazeer, Niki Parmar, Jakob Uszkoreit, Llion Jones,
  Aidan~N Gomez, Łukasz Kaiser, and Illia Polosukhin.
\newblock {Attention is all you need}.
\newblock In \emph{NeurIPS}, pages 5998--6008, 2017.

\bibitem[Walmsley et~al.(2020)Walmsley, Smith, Lintott, Gal, Bamford,
  Dickinson, Fortson, Kruk, Masters, Scarlata, and
  {others}]{Walmsley2020GalaxyLearning}
Mike Walmsley, Lewis Smith, Chris Lintott, Yarin Gal, Steven Bamford, Hugh
  Dickinson, Lucy Fortson, Sandor Kruk, Karen Masters, Claudia Scarlata, and
  {others}.
\newblock {Galaxy Zoo: probabilistic morphology through Bayesian CNNs and
  active learning}.
\newblock \emph{MNRAS}, 491\penalty0 (2):\penalty0 1554--1574, 2020.

\bibitem[Xie and Tu(2015)]{Xie2015Holistically-nestedDetection}
Saining Xie and Zhuowen Tu.
\newblock {Holistically-nested edge detection}.
\newblock In \emph{ICCV}, pages 1395--1403, 2015.

\bibitem[Zhao et~al.(2017)Zhao, Shi, Qi, Wang, and Jia]{Zhao2017PyramidNetwork}
Hengshuang Zhao, Jianping Shi, Xiaojuan Qi, Xiaogang Wang, and Jiaya Jia.
\newblock {Pyramid scene parsing network}.
\newblock In \emph{CVPR}, pages 2881--2890, 2017.

\bibitem[Zhao et~al.(2018)Zhao, Zhang, Liu, Shi, Loy, Lin, and
  Jia]{Zhao2018Psanet:Parsing}
Hengshuang Zhao, Yi~Zhang, Shu Liu, Jianping Shi, Chen~Change Loy, Dahua Lin,
  and Jiaya Jia.
\newblock {Psanet: Point-wise spatial attention network for scene parsing}.
\newblock In \emph{ECCV}, pages 267--283, 2018.

\end{thebibliography}

\end{document}